\documentclass[letterpaper]{article} 
\usepackage[arxiv]{aaai24}  
\usepackage{times}  
\usepackage{helvet}  
\usepackage{courier}  
\usepackage[hyphens]{url}  
\usepackage{graphicx} 
\usepackage{natbib}  
\usepackage{caption} 
\usepackage{algorithm}
\usepackage{algorithmicx}

\usepackage{newfloat}
\usepackage{listings}
\DeclareCaptionStyle{ruled}{labelfont=normalfont,labelsep=colon,strut=off} 
\floatstyle{ruled}
\newfloat{listing}{tb}{lst}{}
\floatname{listing}{Listing}
\usepackage{url}            
\usepackage{booktabs}       
\usepackage{amsfonts}       
\usepackage{nicefrac}       
\usepackage{microtype}      
\usepackage{xcolor}         
\usepackage{amssymb}
\usepackage{amsthm}
\usepackage{amsmath}
\usepackage{algpseudocode}
\usepackage{booktabs}
\usepackage{multirow}
\usepackage{caption}
\usepackage{subcaption}
\newtheorem{theorem}{Theorem}[section]

\newtheorem{lemma}[theorem]{Lemma}
\newtheorem{assumption}[theorem]{Assumption}
\newtheorem{definition}[theorem]{Definition}

\title{Unsupervised Data Generation for Offline Reinforcement Learning: A Perspective from Model}

\author {
    Shuncheng He\textsuperscript{\rm 1},
    Hongchang Zhang\textsuperscript{\rm 1},
    Jianzhun Shao\textsuperscript{\rm 1},
    Yuhang Jiang\textsuperscript{\rm 1},
    Xiangyang Ji\textsuperscript{\rm 1}
}
\affiliations {
    \textsuperscript{\rm 1}Tsinghua University\\
    hesc16@mails.tsinghua.edu.cn
}

\begin{document}

\maketitle

\begin{abstract}
Offline reinforcement learning (RL) recently gains growing interests from RL researchers. However, the performance of offline RL suffers from the out-of-distribution problem, which can be corrected by feedback in online RL. Previous offline RL research focuses on restricting the offline algorithm in in-distribution even in-sample action sampling. In contrast, fewer work pays attention to the influence of the batch data. In this paper, we first build a bridge over the batch data and the performance of offline RL algorithms theoretically, from the perspective of model-based offline RL optimization. We draw a conclusion that, with mild assumptions, the distance between the state-action pair distribution generated by the behavioural policy and the distribution generated by the optimal policy, accounts for the performance gap between the policy learned by model-based offline RL and the optimal policy. Secondly, we reveal that in task-agnostic settings, a series of policies trained by unsupervised RL can minimize the worst-case regret in the performance gap. Inspired by the theoretical conclusions, UDG (Unsupervised Data Generation) is devised to generate data and select proper data for offline training under tasks-agnostic settings. Empirical results demonstrate that UDG can outperform supervised data generation on solving unknown tasks.
\end{abstract}

\section{Introduction}
Reinforcement learning (RL) recently gains significant advances in sequential decision making problems, with applications ranging from the game of Go \cite{silver2016mastering, silver2017mastering}, video games \cite{mnih2015human, hessel2018rainbow}, to autonomous driving \cite{kiran2021deep} and robotic control \cite{zhao2020sim}. However, the costly online trial-and-error process requires numerous samples of interactions with the environment which restricts RL from real world deployment. In the scenarios where online interaction is expensive or unsafe, we have to resort to offline experience \cite{levine2020offline}. However, transplanting RL to offline can provoke disastrous error by falsely overestimating the out-of-distribution samples without correction from environment feedback. Despite recent advances on mitigating bootstrapped error by constraining the policy in data distribution or even in data samples \cite{fujimoto2021minimalist}, offline RL is still limited since they can barely generalize to out-of-distribution areas \cite{yu2020mopo}. The inability of generalization of offline RL will be a serious issue when the batch data deviates from the optimal policy especially under the settings of multi-task, task transfer or task-agnostic. As plenty of research on online RL succeeds \cite{sodhani2021multi, yu2020meta, yu2020gradient, laskin2021urlb, eysenbach2018diversity, sharma2019dynamics}, we hope offline RL can cope with task-agnostic problems either. To this end, how the batch data distributes becomes the primal concern.\par
Recent research empirically shows that diversity in offline data improves performance on task transfer and solving multiple tasks \cite{lambert2022challenges, yarats2022don}. The diverse dataset is obtained from unsupervised RL by competitively training diverse policies \cite{eysenbach2018diversity}, or exploration emphasized pre-training \cite{liu2021aps}, and all of the generated data is fed to offline algorithms. However, these studies barely address the connection between the batch data and the performance of offline RL theoretically. How the diversity of data contributes to solving task-agnostic problems remains unclear.\par
\begin{figure}[ht]
    \centering
    \includegraphics[width=\linewidth]{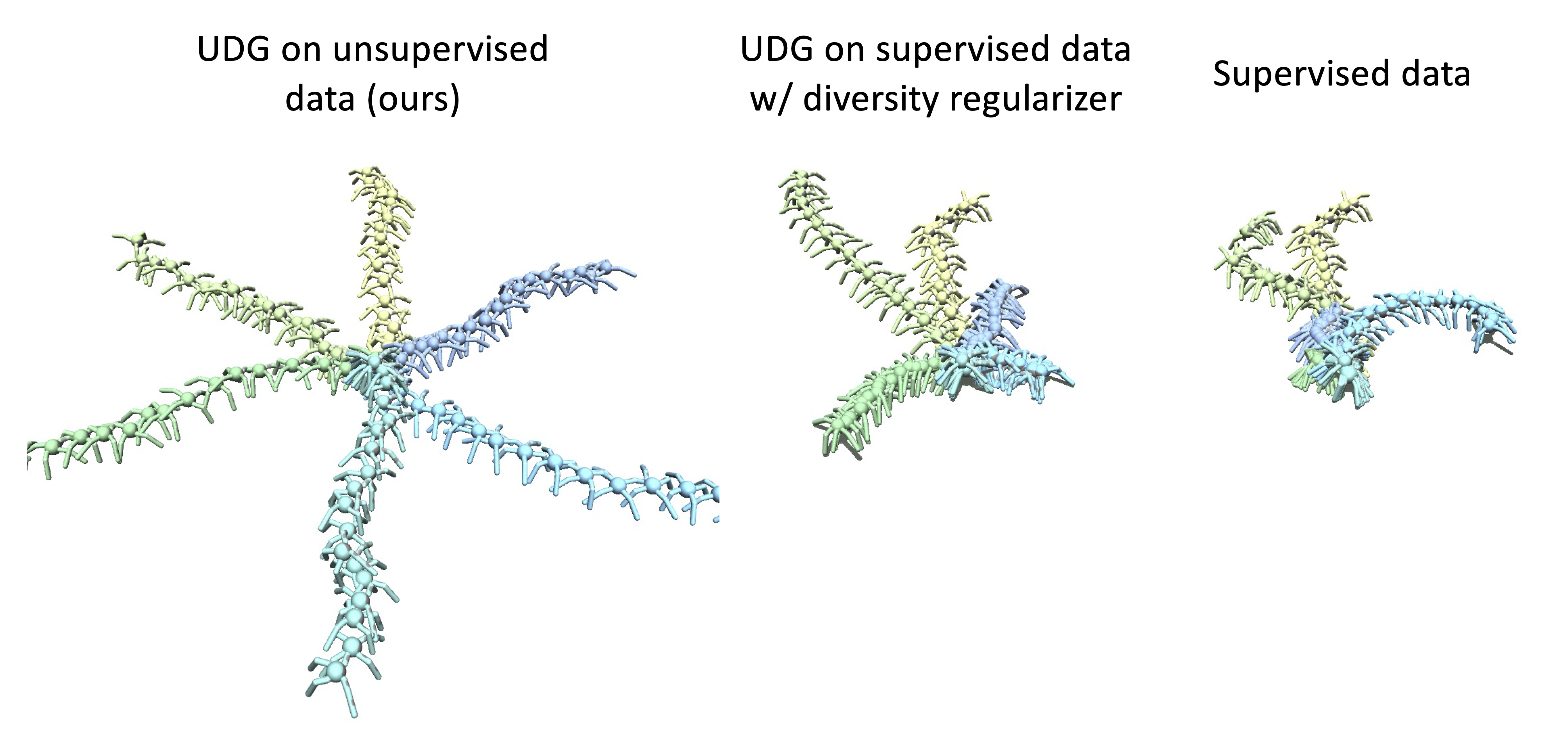}
    \caption{Rendered trajectories of offline trained policies in 6 Ant-Angle tasks. These tasks require the ant to move along 6 different directions. With diverse data buffers generated by unsupervisedly trained policies, our method UDG can solve all the tasks by offline reinforcement learning.}
    \label{fig:render}
\end{figure}
\begin{figure}[ht]
    \centering
    \includegraphics[width=\linewidth]{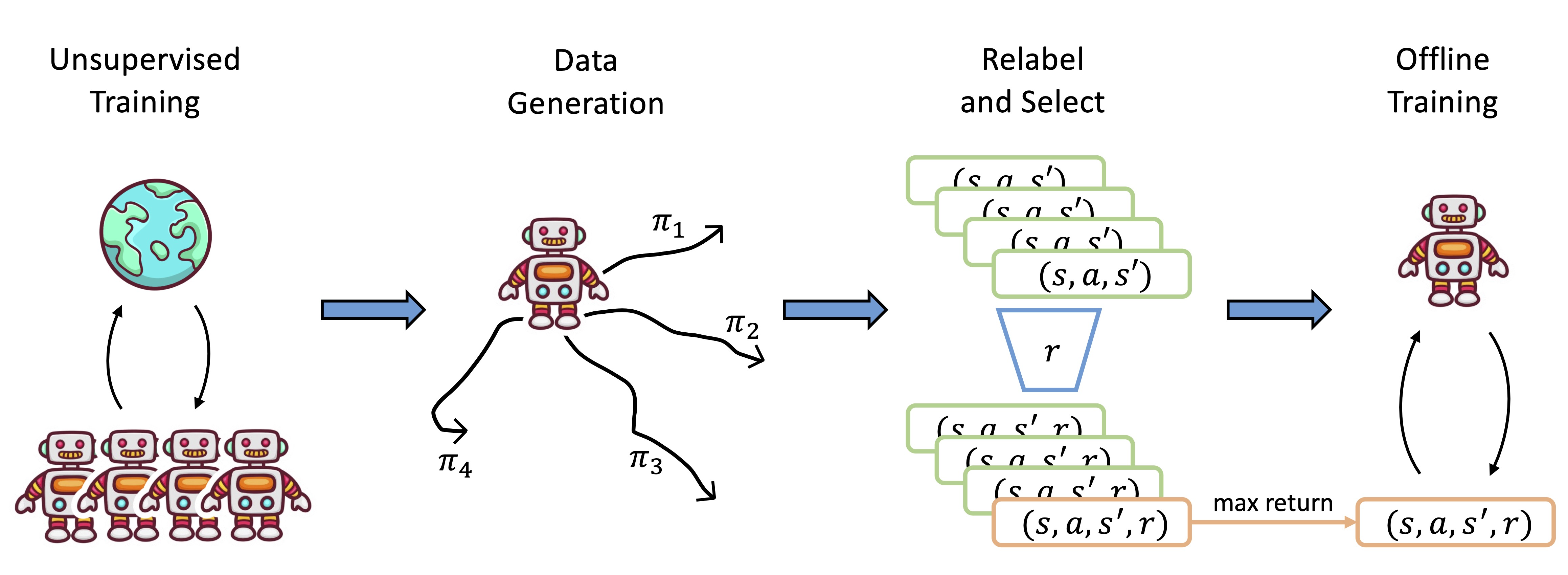}
    \caption{The framework of UDG. First a series of $K$ policies are trained simultaneously with diversity rewards. Second, collect rollout experience $(s,a,s')$ from each policy and construct a corresponding data buffer. Third, relabel the reward in the batch data with a designated reward function, and select the data buffer with the maximal average return. Finally train the agent on the chosen data by offline RL approaches.}
    \label{fig:frame}
\end{figure}

Our study addresses the connection between batch data and performance from a perspective of model based offline optimization MOPO \cite{yu2020mopo}. MOPO establishes a lower bound of the expected return of offline trained policy and reveals that the model prediction error on the optimal data distribution mainly contributes to the performance gap. In this paper, we examine the model prediction error and find the connection between the performance gap and the Wasserstein distance of the batch data distribution from the optimal distribution. We conclude that the offline trained policy will have higher return close to the optimal policy if the behavioural distribution is closer to the optimal distribution. We discover that, in task-agnostic scenarios, unsupervised RL methods which propel the policies far away from each other, approximately optimize the minimal regret to the optimal policy. Based on these theoretical analysis, we propose a framework named unsupervised data generation (UDG) as illustrated in Figure \ref{fig:frame}. In UDG, a series of policies are trained with diversity rewards. They are used to generate batch data stored in different buffers. Before the offline training stage, the buffers are relabeled with given reward function corresponding to the task, and the buffer with highest return is sent to the offline algorithm.\par

The contributions in this work are three-fold. First, to our best knowledge, we are the first to establish a theoretical bond between the behavioural batch data and the performance of offline RL algorithms on Lipschitiz continuous environments. Second, we present an objective of minimal worst-case regret for data generation on task-agnostic problems. Third, we propose a new framework UDG for unsupervised offline RL and evaluate UDG on locomotive environments. Empirical results on locomotive tasks like Ant-Angle and Cheetah-Jump show that UDG outperforms conventional offline RL with random or supervised data. Further experiments validate the soundness of our theoretical findings.

\section{Related work}
\textbf{Offline RL}. 
Online reinforcement requires enormous samples and makes RL less feasible in many real-world tasks \cite{levine2020offline}. Therefore, learning from batch data is a key route to overcome high sample complexity. However, vanilla online RL algorithms face the out-of-distribution problem that the Q function may output falsely high values on samples not in data distribution. To mitigate this issue, model-free offline RL methods cope with the out-of-distribution problem from two aspects, constraining the learned policy on the support of batch data \cite{fujimoto2019off, kumar2019stabilizing, wu2019behavior, peng2019advantage, siegel2020keep, cheng2022adversarially, rezaeifar2022offline, zhang2022state, fujimoto2021minimalist}, and suppressing Q values on out-of-distribution area \cite{agarwal2019striving, kumar2020conservative}. To generalize beyond the batch data, model-based offline RL methods employ transition models to produce extra samples for offline learning \cite{chen2021offline, kidambi2020morel, yu2020mopo, matsushima2020deployment}. These methods can naturally generalize to areas where the model is accurate \cite{janner2019trust}. Our theoretical work originates from MOPO \cite{yu2020mopo}, a model-based offline algorithm by adding an uncertainty penalty to avoid unexpected exploitation when model is inaccurate. MOPO derives a performance lower bound w.r.t. model prediction error. 
In contrast, we investigate the lower bound w.r.t. data to show the connection between data distribution and performance. On previous theoretical analysis on offline RL, the data coverage condition is crucial for stable, convergent offline learning \cite{wang2020statistical, chen2019information}. In this paper, we focus on Lipschitz continuous environments which are common in locomotive tasks such as HalfCheetah, Ant in MuJoCo \cite{todorov2012physics}. By investigating into the transition with Lipschitz geometry, our analysis makes no assumptions on data coverage.\par
\textbf{Unsupervised RL}. 
Reinforcement learning heavily relies on the reward feedback of the environment for a specific tasks. However, recent research on unsupervised RL demonstrate training policies without extrinsic rewards enables the agent to adapt to general tasks \cite{laskin2021urlb, eysenbach2018diversity}. Without supervision from extrinsic rewards, unsupervised RL methods can either be driven by curiosity/novelty \cite{pathak2017curiosity, pathak2019self, burda2018exploration}, maximum coverage of the state space \cite{liu2021behavior, campos2020explore, yarats2021reinforcement}, and diversity of a series of policies \cite{florensa2017stochastic, lee2019efficient, eysenbach2018diversity, sharma2019dynamics, he2022wasserstein, liu2021aps, strouse2021learning, kim2021unsupervised}. These methods all provide a pseudo reward derived from their own criteria. Our work employs a diverse series of policies to generate batch data for offline learning under task-agnostic settings. Therefore we utilize an unsupervised training paradigm in alignment with DIAYN \cite{eysenbach2018diversity}, DADS \cite{sharma2019dynamics}, WURL \cite{he2022wasserstein}, and choose WURL as base algorithm in accordance with our theoretical analysis.\par
\textbf{Offline dataset}. 
D4RL \cite{fu2020d4rl} and RL Unplugged \cite{gulcehre2020rl} are most commonly used offline RL benchmarks. The datasets in these benchmarks consist of replay buffers during training, rollout samples generated by a policy of a specific level, or samples mixed from different policies. Apart from benchmark datasets, exploratory data gains growing interest \cite{wang2022offline}. Explore2Offline \cite{lambert2022challenges} and ExORL \cite{yarats2022don} both investigate into the role of batch data and construct a more diverse dataset with unsupervised RL algorithms for task generalization. In addition, extensive experiments in ExORL empirically show exploratory data can improve the performance of offline RL algorithms and even TD3 \cite{fujimoto2018addressing}. However, neither of these methods have theoretical analysis on the connection between data diversity and offline performance. Our work completes the picture of how diverse data improves offline RL performance. And we point out that data selection before offline training has considerable influence on performance, which is not addressed in previous work.
\section{Preliminaries}
A Markov decision process (MDP) is formalized as $M=(\mathcal{S},\mathcal{A},T,r,p_{0},\gamma)$, where $\mathcal{S}$ denotes the state space while $\mathcal{A}$ denotes the action space, $T(s'|s,a)$ the transition dynamics, $r(s,a)$ the reward function, $p_{0}$ the initial state distribution and $\gamma\in[0,1)$ the discounted factor. RL aims to solve the MDP by finding a policy $\pi(a|s)$ maximizing the expected accumulated discounted return $\eta_{M}(\pi):=\mathbb{E}_{\pi,T,p_{0}}[\sum_{t=0}^{\infty}\gamma^{t}r(s_{t},a_{t})]$. The state value function $V_{M}^{\pi}(s_{t}):=\mathbb{E}_{\pi,T}[\sum_{k=0}^{\infty}\gamma^{k}r(s_{t+k},a_{t+k})]$ provides an expectation of discounted future return from $s_{t}$ under policy $\pi$ and MDP $M$. Let $\mathbb{P}_{T,t}^{\pi}(s)$ be the probability of being in state $s$ at step $t$ when acting with policy $\pi$. The discounted occupancy measure of $\pi$ under dynamics $T$ is denoted by $\rho_{T}^{\pi}(s,a):=\frac{1}{c}\pi(a|s)\sum_{t=0}^{\infty}\gamma^{t}\mathbb{P}_{T,t}^{\pi}(s)$ where $c=1/(1-\gamma)$ is the normalization constant. Likewise, $\rho_{T}^{\pi}(s):=\frac{1}{c}\sum_{t=0}^{\infty}\gamma^{t}\mathbb{P}_{T,t}^{\pi}(s)$ is the state occupancy distribution. The expected accumulated discounted return can be rewritten as $\eta_{M}^{\pi}=c\mathbb{E}_{\rho_{T}^{\pi}}[r(s,a)]$. $\hat{\rho}_{T}^{\pi}=\frac{1}{K}\sum_{i=1}^{K}\delta(s_{i},a_{i})$ denotes the empirical distribution of $\rho_{T}^{\pi}$ based on $K$ samples sampled from $\pi$ under transition dynamics $T$. $\delta(\cdot)$ denotes Dirac distribution.\par
Model-based RL approaches learn an estimated model $\hat{T}$ from interaction experience, which defines a model MDP $\hat{M}=(\mathcal{S},\mathcal{A},\hat{T},r,p_{0},\gamma)$. Similarly we have the expected return under the learned dynamics $\eta_{\hat{M}}(\pi)=c\mathbb{E}_{\rho_{\hat{T}}^{\pi}}[r(s,a)]$.\par
In offline settings, the RL algorithm optimizes the policy solely on a fixed dataset $\mathcal{D}_{\beta}={(s,a,r,s')}$ generated by the behavioural policy $\pi^{\beta}$. $\pi^{\beta}$ can be one policy or a mixture of policies. Note that offline RL algorithms cannot interact with the environment or produce extra samples. In model-based offline RL, the algorithm first learn a transition model $\hat{T}$ from the batch data $\mathcal{D}_{\beta}$. At the training stage, the algorithm executes $k$-step rollout using the estimated model from the state sample from $\mathcal{D}_{\beta}$. The generated data are added to another buffer $\mathcal{D}_{m}$. Both data buffers are used in offline policy optimization.\par
\section{UDG: Unsupervised Data Generation}
In order to find the connenction between performance and data, we first review key propositions in MOPO \cite{yu2020mopo}. As discussed before, offline RL faces a dilemma of out-of-distribution samples and lack of exploration. Model-based RL like MBPO \cite{janner2019trust} can naturally extend to the regions where the model predicts as well as the true dynamics. However, when the model is inaccurate, the algorithm may exploit the falsely high return regions, resulting in inferior test performance in true dynamics. MOPO first derives the performance lower bound represented by the model error, and then addresses the risk-return trade-off by incorporating the penalty represented by the error of the estimated dynamics into the reward of offline policy optimization.\par
We briefly summarize the derivation of the performance lower bound. First we introduce the telescoping lemma:
\begin{lemma}\label{lem1}
Let $M$ and $\hat{M}$ be two MDPs with the same reward function $r$, but different dynamics $T$ and $\hat{T}$ respectively. Denote $G_{\hat{M}}^{\pi}(s,a):=\mathbb{E}_{s'\sim\hat{T}(s,a)}[V_{M}^{\pi}(s')]-\mathbb{E}_{s'\sim T(s,a)}[V_{M}^{\pi}(s')]$. Then
\begin{equation}
\eta_{\hat{M}}(\pi)-\eta_{M}(\pi)=c\gamma\mathbb{E}_{(s,a)\sim\rho_{\hat{T}}^{\pi}}\left[G_{\hat{M}}^{\pi}(s,a)\right].
\end{equation}
\end{lemma}
If we have mild constraints on the value function $V_{M}^{\pi}\in\mathcal{F}$ where $\mathcal{F}$ is a bounded function class under a specific metric, then we can bound the gap $G_{\hat{M}}^{\pi}(s,a)$ with model error measured by corresponding integral probability measure (IPM) $d_{\mathcal{F}}$ \cite{muller1997integral},
\begin{equation}
\begin{split}
  |G_{\hat{M}}^{\pi}(s,a)|&\leq\sup_{f\in\mathcal{F}}\left|\mathbb{E}_{s'\sim\hat{T}(s,a)}[f(s')]-\mathbb{E}_{s'\sim T(s,a)}[f(s')]\right|\\
  &=d_{\mathcal{F}}(\hat{T}(s,a),T(s,a)).  
\end{split}
\end{equation}\par 
Since we cannot access the true model $T$ in most cases, MOPO adopts an admissible error estimator $u:\mathcal{S}\times\mathcal{A}\rightarrow\mathbb{R}$ for $\hat{T}$, and have an assumption that for all $s\in\mathcal{S}. a\in\mathcal{A}$, $d_{\mathcal{F}}(\hat{T}(s,a),T(s,a))\leq u(s,a)$. An uncertainty-penalized MDP $\tilde{M}=(\mathcal{S},\mathcal{A},\hat{T},\tilde{r},p_{0},\gamma)$ is defined given the error estimator $u$, with the reward $\tilde{r}(s,a):=r(s,a)-\gamma u(s,a)$ penalized by model error.\par
By optimizing the policy in the uncertainty-penalized MDP $\tilde{M}$, MOPO has a following performance lower bound,
\begin{theorem}[MOPO]\label{thm1}
Given $\hat{\pi}=\arg\max_{\pi}\eta_{\tilde{M}}(\pi)$ and $\epsilon_{u}(\pi):=c\mathbb{E}_{(s,a)\sim\rho_{\hat{T}}^{\pi}}[u(s,a)]$, the expected discounted return of $\hat{\pi}$ satisfies
\begin{equation}
\eta_{M}(\hat{\pi})\geq\sup_{\pi}\left\{\eta_{M}(\pi)-2\gamma\epsilon_{u}(\pi)\right\}.
\end{equation}
\end{theorem}
The theorem \ref{thm1} reveals the optimality gap between $\pi^{*}$ and $\hat{\pi}$. Immediately we have $\eta_{M}(\hat{\pi})\geq\eta_{M}(\pi^{*})-2\gamma\epsilon_{u}(\pi^{*})$. This corollary indicates if the model error is small on the $(s,a)$ occupancy distribution under the optimal policy $\pi^{*}$ and dynamics $\hat{T}$, the optimality gap will be small.
In order to find the deep connection between the batch data and the performance gap of model-based offline RL algorithms, in the following section, we directly analyze the model prediction deviation $d_{\mathcal{F}}(\hat{T}(s,a),T(s,a))$ instead of the error estimator $u(s,a)$.\par
\subsection{The connection between batch data and offline RL performance}
Before presenting a lower bound of $\eta_{M}(\hat{\pi})$, here we make a few assumptions to simplify the proof. Some of these assumptions can be loosen and do not change the conclusion of the main result. The generalization of the theoretical analysis is discussed in Appendix A.\par
\begin{definition}
Given a bounded subset $\mathcal{K}$ in the corresponding $d$-dimension Euclidean space $\mathbb{R}^{d}$. The diameter $B_{\mathcal{K}}$ of set $\mathcal{K}$ is defined as the minimum value of $B$ such that there exists $k_{0}\in\mathbb{R}^{d}$, for all $k\in\mathcal{K}$, $\|k_{0}-k\|\leq B$.
\end{definition}
\begin{assumption}\label{ass1}
The state space $\mathcal{S}$, the action space $\mathcal{A}$ are both bounded subsets of corresponding Euclidean spaces, with diameter $B_{\mathcal{A}}\ll B_{\mathcal{S}}$. The state transition function $T(s'|s,a)$ is deterministic and continuous.
\end{assumption}
\begin{assumption}\label{ass2}
The state transition function $T(s'|s,a)$ is $L_{T}$-Lipschitz. For any $\pi$ the value function $V^{\pi}(s)$ is $L_{r}$-Lipschitz.
\end{assumption}
 As a consequence, given two state-action pairs $(s_{1},a_{1}),(s_{2},a_{2})$, the next-state deviation under transition $T$ is upper bounded by
\begin{equation}
\|T(s_{1},a_{1})-T(s_{2},a_{2})\|\leq L_{T}\|(s_{1},a_{1})-(s_{2},a_{2})\|.
\end{equation}
\begin{assumption}\label{ass3}
The prediction model $\hat{T}(s'|s,a)$ is a non-parametric transition model, which means the model outputs the next state prediction by searching the nearest entry.
\end{assumption}
Formally speaking, $\hat{T}$ has an episodic memory storing all input experience $\mathcal{D}_{\text{memory}}=\{(s_{i},a_{i},s'_{i},r_{i})\}_{i=1}^{K}$ \cite{pritzel2017neural}. When feeding $\hat{T}$ a query $(s,a)$, the model returns $\hat{T}(s,a)=s'_{k}$ where $k=\arg\min_{i}\|(s,a)-(s_{i},a_{i})\|$. Assumption \ref{ass3} implies $\hat{T}(s,a)$ is a deterministic function. Therefore combined with these two assumptions \ref{ass1}, \ref{ass2}, the gap $G_{\hat{M}}^{\pi}(s,a)$ defined in Lemma \ref{lem1} is then bounded by
\begin{equation}\label{eqn5}
\begin{split}
    |G_{\hat{M}}^{\pi}(s,a)|&\leq L_{r}W_{1}(\hat{T}(s,a),T(s,a))\\
    &=L_{r}\|\hat{T}(s,a)-T(s,a)\|,
\end{split}
\end{equation}
where $W_{1}$ is the 1-Wasserstein distance w.r.t. the Euclidean metric.
\begin{assumption}\label{ass4}
$\rho_{T}^{\pi^{\beta}}$ have a bounded support. The diameter of the support of distribution $\rho_{T}^{\pi^{\beta}}$ is denoted as $B_{\pi^{\beta}}$.
\end{assumption}
Since the dataset size is out of our concern, we suppose the batch data is sufficient, such that $\hat{\rho}_{T}^{\pi^{\beta}}\approx\rho_{T}^{\pi^{\beta}}\approx\rho_{\hat{T}}^{\pi^{\beta}}$. For conciseness, we use $\rho_{T}^{\pi^{\beta}}$ in the following statements.
\begin{theorem}\label{thm2}
Given $\hat{\pi}=\arg\max_{\pi}\eta_{\tilde{M}}(\pi)$, the expected discounted return of $\hat{\pi}$ satisfies
\begin{equation}\label{eqn6}
\begin{split}
\eta_{M}(\hat{\pi})&\geq\eta_{M}(\pi^{*})-C(W_{1}(\rho_{T}^{\pi^{*}},\rho_{\hat{T}}^{\pi^{*}})+W_{1}(\rho_{T}^{\pi^{\beta}},\rho_{T}^{\pi^{*}}))\\
&\geq\eta_{M}(\pi^{*})-2C(W_{1}(\rho_{T}^{\pi^{\beta}},\rho_{T}^{\pi^{*}})+B_{\pi^{\beta}}+B_{\mathcal{A}}),
\end{split}
\end{equation}
where $C=2c\gamma L_{r}L_{T}$ is the constant related to the assumptions. If the batch data is collected from $N$ different policies $\pi_{1}^{\beta},\dots,\pi_{N}^{\beta}$, a tighter bound is obtained, where $\rho_{T}^{\pi^{\beta}}$ denotes the mixture of distribution $\rho_{T}^{\pi_{i}^{\beta}},\dots,\rho_{T}^{\pi_{N}^{\beta}}$,
\begin{equation}\label{eqn7}
\begin{split}
   \eta_{M}(\hat{\pi})&\geq\eta_{M}(\pi^{*})-C(W_{1}(\rho_{T}^{\pi^{\beta}},\rho_{T}^{\pi^{*}})\\
   &+\min_{i}W_{1}(\rho_{T}^{\pi_{i}^{\beta}},\rho_{T}^{*})+2B_{\pi^{\beta}}+2B_{\mathcal{A}}). 
\end{split}
\end{equation}
\end{theorem}
The distance term $D_{1}:=W_{1}(\rho_{T}^{\pi^{*}}(s,a),\rho_{\hat{T}}^{\pi^{*}}(s,a))$ in the first line in Equation \ref{eqn6} is quite hard to estimate. However, we notice that the prediction model only outputs states in the episodic memory $\mathcal{D}_{\text{memory}}$ which implies $\text{supp}(\rho_{\hat{T}}^{\pi^{*}}(s))\subseteq\text{supp}(\rho_{T}^{\pi^{\beta}}(s))$. We can naturally suppose that $\rho_{\hat{T}}^{\pi^{*}}(s,a)$ will not be too distinct from $\hat{\rho}_{T}^{\pi^{\beta}}(s,a)$. Therefore we can assume that $D_{1}\approx D_{2}:=W_{1}(\rho_{T}^{\pi^{\beta}}(s,a),\rho_{T}^{\pi^{*}}(s,a))$, leading to an approximate lower bound free of $B_{\pi^{\beta}}$ and $B_{\mathcal{A}}$
\begin{equation}\label{eqn8}
\eta_{M}(\hat{\pi})\geq\eta_{M}(\pi^{*})-2C(W_{1}(\rho_{T}^{\pi^{\beta}},\rho_{T}^{\pi^{*}})).
\end{equation}
For the data compounded by a mixture of policies, the approximate lower bound is
\begin{equation}\label{eqn9}
\eta_{M}(\hat{\pi})\geq\eta_{M}(\pi^{*})-C(W_{1}(\rho_{T}^{\pi^{\beta}},\rho_{T}^{\pi^{*}})+\min_{i}W_{1}(\rho_{T}^{\pi_{i}^{\beta}},\rho_{T}^{*})).
\end{equation}
The detailed proof of Theorem \ref{thm2} is presented in Appendix A. The main idea is to show the performance gap $\mathbb{E}_{(s,a)\sim\rho_{\hat{T}}^{\pi^{*}}}|G_{\hat{M}}^{\pi^{*}}(s,a)|$ can be bounded by the distance between $\rho_{\hat{T}}^{\pi^{*}}$ and $\rho_{\hat{T}}^{\pi^{\beta}}$. The remaining part of proof utilizes triangle inequality to split the distance into two terms and then applies the assumptions to yield Equation \ref{eqn6}.
\paragraph{Interpretation:} Theorem \ref{thm2} and Equation \ref{eqn8} suggest that the gap relies on $\pi^{\beta}$ and $\pi^{*}$. We denote the gap as $\mathcal{L}(\pi^{\beta},\pi^{*})$ such that $\eta_{M}(\hat{\pi})\geq\eta_{M}(\pi^{*})-\mathcal{L}(\pi^{\beta},\pi^{*})$. When the occupancy distribution of $\pi^{\beta}$ is closer to the occupancy distribution of the optimal policy $\pi^{*}$, the return of the policy optimized by MOPO will be closer to the optimal. Especially when $\pi^{\beta}=\pi^{*}$, MOPO can reach the optimal return. Theorem \ref{thm2} concentrates on the gap between $\pi^{\beta}$ and $\pi^{*}$. However, since the derivation does not involve the optimality of $\pi^{*}$, the inequality \ref{eqn6} holds true for any other policy $\pi$ instead of the optimal policy $\pi^{*}$. By substituting $\pi^{*}$ with $\pi^{\beta}$, we will obtain $\eta_{M}(\hat{\pi})\geq\eta_{M}(\pi^{\beta})$, which means the performance of the learned policy will perform no worse than the behavioral policy. This conclusion is consistent with the theoretical analysis in MOPO.\par
The second line of Equation \ref{eqn6} indicates a wider range of $\rho_{T}^{\pi^{\beta}}$ may enlarge the performance gap. This issue is mainly determined by the relation between $\rho_{\hat{T}}^{\pi^{*}}(s,a)$ and $\rho_{T}^{\pi^{\beta}}(s,a)$. In general cases, $\pi^{*}(a|s)$ will output actions that lead the next states closer to the optimal occupancy distribution. As a consequence, $\rho_{\hat{T}}^{\pi^{*}}(s,a)$ may be closer to $\rho_{T}^{\pi^{*}}(s,a)$ than $\rho_{T}^{\pi^{\beta}}(s,a)$. Therefore $D_{1}$ will be smaller than $D_{2}$. Nevertheless, $D_{1}>D_{2}$ is still possible under some non-smooth dynamics or multi-modal situations. As a result, a broader distribution of $\rho_{T}^{\pi^{\beta}}(s,a)$ may impair MOPO performance.\par
\subsection{The minimal worst-case regret approach}
There are many cases where the optimal policy and the corresponding experience data is inaccessible for offline learning, e.g., (1) the reward function is unknown or partly unknown at the stage of data generation; (2) the batch data is prepared for multiple tasks with various reward functions; (3) training to optimal is expensive at the stage of data generation.  Previous work in online RL suggests the diversity of policies plays the crucial role \cite{eysenbach2018diversity}. Especially in offline RL, where exploration is not feasible during training, the diversity of batch data should not be ignored.\par
Suppose we can train a series of $N$ policies simultaneously without any external reward. Our goal is to improve the diversity of the experience collected by policies $\{\pi_{i}\}_{i=1}^{N}$, such that there is at least one subset of the experience will be close enough to the optimal policy determined by the lately designated reward function at the offline training stage. Combined with Theorem \ref{thm2}, this objective can be formulated by
\begin{equation}\label{eqn10}
\min_{\pi_{1},\dots,\pi_{N}\in\Pi}\max_{\pi^{*}\in\Pi}\min_{i}\mathcal{L}(\pi^{i},\pi^{*}).
\end{equation}
The inner $\min$ term $\text{REGRET}(\{\pi_{i}\}_{i=1}^{N},\pi^{*}):=\min_{i}\mathcal{L}(\pi_{i},\pi^{*})$ represents the regret of the series of policies confronting the true reward function and its associate optimal policy. Since we are able to choose one policy in the series after informed of the optimal policy, we take the minimum as the regret. The $\max$ operator in the middle depicts the worst-case regret if any policy in the feasible policy set $\Pi$ has the possibility to be the optimal one. The outer $\min$ means the goal of optimizing $\{\pi_{i}\}_{i=1}^{N}$ is to minimize the worst-case regret. If the approximate lower bound is considered in Equation \ref{eqn8}, the objective is equivalent to 
\begin{equation}\label{eqn11}
\min_{\pi_{1},\dots,\pi_{N}\in\Pi}\max_{\pi^{*}\in\Pi}\min_{i}W_{1}(\rho_{T}^{\pi_{i}},\rho_{T}^{\pi^{*}}).
\end{equation}
Directly optimizing a series of policies according to the minimax objective in Equation \ref{eqn11} inevitably requires adversarial training. Previous practice suggests an adversarial policy should be introduced to maximize $\text{REGRET}(\{\pi_{i}\}_{i=1}^{N},\pi^{*})$, playing the role of the unknown optimal policy. The adversarial manner of training brings us two main concerns. (1) Adversarial training may incur instability and require much more steps to converge \cite{arjovsky2017towards, arjovsky2017wasserstein}; (2) The regret only provides supervision signals to the policy nearest to $\pi^{*}$, which leads to low efficiency in optimization.\par
Eysenbach et al. \cite{eysenbach2021information} proposed similar objective regarding unsupervised reinforcement learning. Under the assumptions of finite and discrete state space and, the quantity of policies $N$ should cover the number of distinct states in the state space $|\mathcal{S}|$, the minimal worst-case regret objective is equivalent to the maximal discriminability objective. Likewise, we propose a surrogate objective
\begin{equation}\label{eqn12}
\max_{\pi_{1},\dots,\pi_{N}\in\Pi}\min_{i\neq j}W_{1}(\rho_{T}^{\pi_{i}},\rho_{T}^{\pi_{j}}).
\end{equation}
The surrogate objective shares the same spirit with WURL \cite{he2022wasserstein}. Both of them encourage diversity of a series of policies w.r.t. Wasserstein distance in the probability space of state occupancy. Although the optimal solution of $\{\pi_{i}\}_{i=1}^{N}$ does not match the optimal solution in Equation \ref{eqn11} in general situations, both of them represent a kind of diversity. The relation between two objectives equals to the relation between finite covering and finite packing problems, which are notoriously difficult to analyze even in low-dimension, convex settings \cite{boroczky2004finite, toth2017handbook}. Nevertheless, we assume the gap will be small and the surrogate objective will be a satisfactory proxy of Equation \ref{eqn11} as previous literature does in the application of computational graphics \cite{schlomer2011farthest, chen2004optimal}. Refer to Appendix B for more details.\par
\subsection{Practical implementation}
To achieve diversity, practical algorithms assign a pseudo reward $\tilde{r}_{i}$ to policy $\pi_{i}$. The pseudo reward usually indicates the ``novelty'' of the current policy w.r.t. all other policies. Similar to WURL, we adopt pseudo reward $\tilde{r}_{i}:=\min_{j\neq i}W_{1}(\rho_{T}^{\pi_{j}},\rho_{T}^{\pi_{i}})$ which is the minimum distance from all other policies. We compute the Wasserstein distance using amortized primal form estimation in consistent with WURL \cite{he2022wasserstein}.\par
In semi-supervised cases, only part of reward function is known. For example, in Mujoco simulation environments in OpenAI Gym \cite{brockman2016openai}, the complete reward function is composed of a reward related to the task, and general rewards related to agent's health, control cost, safety constraint, etc. We can train the series of policies with a partial reward and a pseudo reward simultaneously by reweighting two rewards with a hyperparameter $\lambda$. Moreover, the complete reward and the diversity-induced pseudo reward can be combined to train a diverse series of policies for generalization purposes.\par
The policies are trained with Soft Actor-Critic method \cite{haarnoja2018soft}. The network model of the actors are stored to generate experience $\mathcal{D}_{1},\dots,\mathcal{D}_{K}$ for offline RL. When a different reward function is used at the offline training stage. The reward will be relabeled with $r(s,a)$. At the offline learning stage, we choose the best buffer and feed it to MOPO. The overall algorithm is illustrated in Figure \ref{fig:frame} and formally described in Algorithm \ref{alg}.
\begin{algorithm}
\caption{Unsupervised data generation for offline RL in task-agnostic settings}
\label{alg}
\textbf{Require:} $K$ policies $\pi_{1},\dots,\pi_{K}$. $K$ empty buffers $\mathcal{D}_{1},\dots,\mathcal{D}_{K}=\{\}$. Maximum buffer size $N$.
\begin{algorithmic}[1]
\State Train $\pi_{i},i=1,\dots,K$ with SAC w.r.t. diversity rewards $\tilde{r}_{i}:=\min_{j\neq i}W_{1}(\rho_{T}^{\pi_{j}},\rho_{T}^{\pi_{i}})$.
\State Let each $\pi_{i}$ interacts with environment for $N$ steps and fill $\mathcal{D}_{i}$ with transitions $(s,a,s')$.
\State Acquire the task and relabel all transitions in $\mathcal{D}_{1},\dots,\mathcal{D}_{K}$ with given $r(s,a)$.
\State Evaluate each buffer and calculate the average return $\bar{G}_{i},i=1,\dots,K$.
\State Select the buffer $\mathcal{D}_{k}$ where $k=\arg\max_{i}\bar{G}_{i}$
\State Train the policy by MOPO with $\mathcal{D}_{k}$.
\end{algorithmic}
\end{algorithm}
\section{Experiments}
\begin{figure}[ht]
    \begin{center}
        \includegraphics[width=0.85\linewidth]{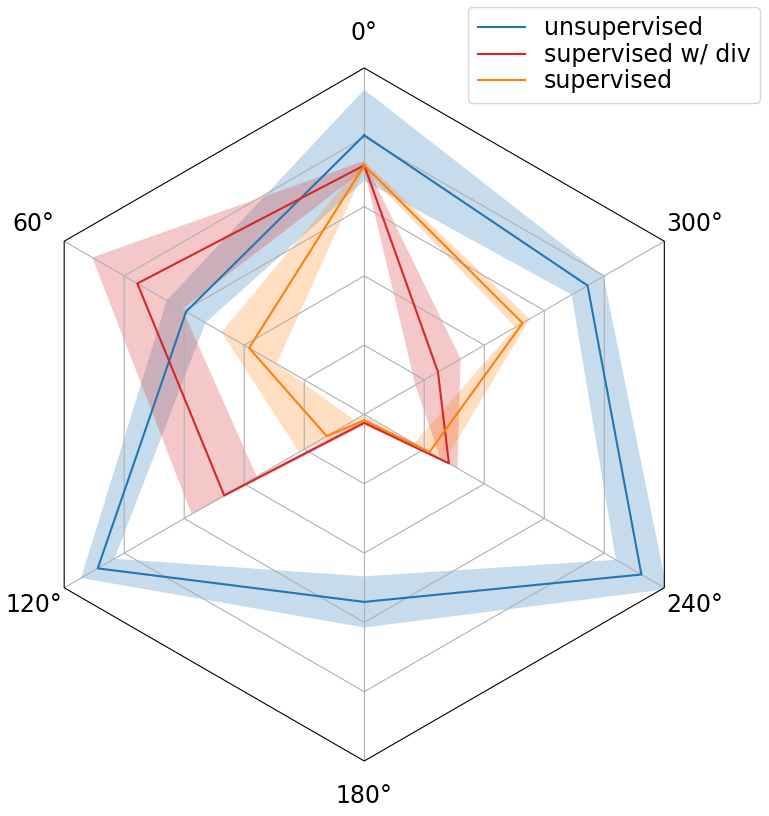}
    \end{center}
    \caption{Results on Ant-Angle tasks. The data buffers of all three methods are evaluated by MOPO with 3 random seeds. The darker lines represent the average evaluation return. The lighter areas depict the standard deviation of the return.}
    \label{fig:radar}
\end{figure}
Based on our framework of UDG in Figure \ref{fig:frame}, we conduct experiments on two locomotive environments requiring the agent to solve a series of tasks with different reward functions at the offline stage. Both of the tasks are re-designed Mujoco environments. Ant-Angle is a task modified from Ant environment. In Ant-Angle, the agent should actuate the ant to move from the initial position to a specific direction on the x-y plane. The agent is rewarded by the inner product of the moving direction and the desired direction. The goal is to construct a dataset while the desired direction is unknown until the offline stage. Cheetah-Jump is another task for evaluation, modified from HalfCheetah environment. The reward in Cheetah-Jump consists of three parts, control cost, velocity reward, and jumping reward. At the data generation stage, the agent can only have access to the control cost and the velocity reward for reducing energy cost of actuators and moving the cheetah forward. The jumping reward is added in offline training, by calculating the positive offset of the cheetah on the z axis. Likewise, a crawling reward can be added to encourage the cheetah to lower the body while moving forward.\par
Our experiments mainly focus on two aspects: (1) How does UDG framework perform on the two challenging tasks, Ant-Angle and Cheetah-Jump? (2) Can experimental results match the findings in the theoretical analysis?
\subsection{Evaluation on task-agnostic settings}
To answer question (1), we construct three types of data buffer. The first is generated by unsupervisedly trained policies with the objective in Equation \ref{eqn12}. The second is created by one supervisedly trained policy that maximizes a specific task reward. The third is the combination of two, which means the polices are trained with both task reward and diversity reward, reweighted by $\lambda$. We call the combination as supervised training with diversity regularizer. Note that the supervised method contains one data buffer. Another two methods have a series of 10 buffers w.r.t. 10 polices and only one buffer is selected during offline training.\par
\begin{figure}
    \centering
    \includegraphics[width=0.95\linewidth]{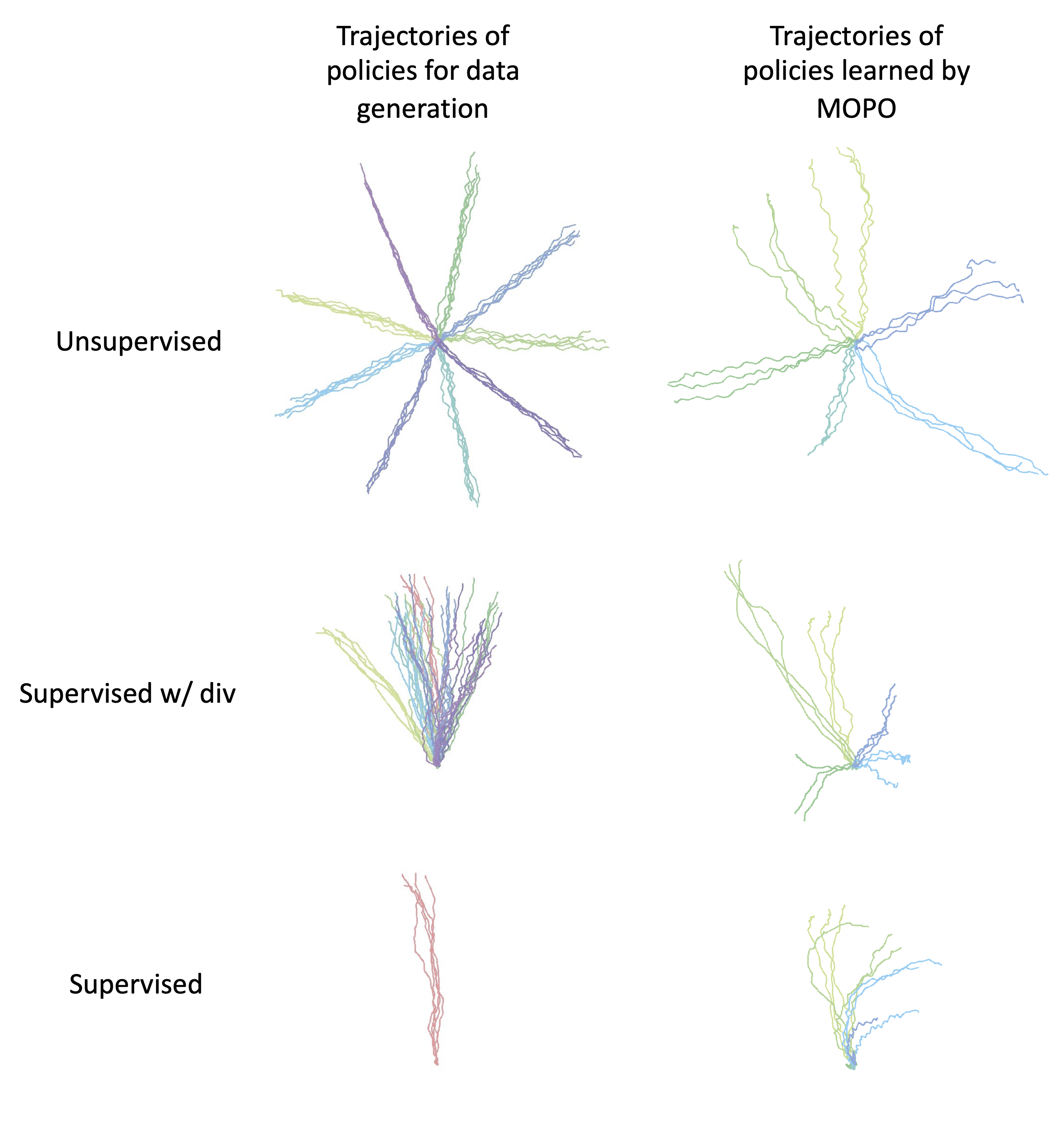}
    \caption{The trajectories of ant on the x-y plane. The lower left figure demonstrates UDG can solve all 6 offline tasks on account of the diversity of unsupervised learned polices. The trajectories in the lower right figure shows the policy learned by MOPO cannot generalize to the directions that largely deviate from $0^{\circ}$.}
    \label{fig:xytraj}
\end{figure}

We evaluate three kinds of data buffers on Ant-Angle. The supervised policy and the supervised policies with diversity regularizer are provided with reward to move in direction $0^{\circ}$, the upper direction in Figure \ref{fig:xytraj}. The diversity reward is calculated on ant position instead of the whole state space, considering that the dimension of state space is extremely high. We evaluate three approaches on 6 offline tasks of moving along the directions of $0^{\circ}$, $60^{\circ}$, $120^{\circ}$, $180^{\circ}$, $240^{\circ}$ and $300^{\circ}$.
As Figure \ref{fig:xytraj} shows, the policies trained with unsupervised RL are evenly distributed over the x-y plane. Therefore in the downstream offline tasks, no matter what direction the task needs, there exists at least one policy that could obtain relatively high reward. The trajectories of policies trained by MOPO confirm that UDG can handle all 6 tasks. Meanwhile the policy trained to move in the direction $0^{\circ}$ generates narrow data and MOPO cannot perform well on other directions. The policies trained with combined reward have wider range of data distribution. Especially, an policy deviates to circa $30^{\circ}$ and consequently the policy trained by MOPO acquires high reward in the $60^{\circ}$ task.\par
\begin{table}[ht]
    \setlength{\abovecaptionskip}{-0.cm}
    \setlength{\belowcaptionskip}{-0.3cm}
    \centering
    \begin{tabular}{cccc}
    \toprule
    Task     &   $c_{z}$     &   random      &   diverse     \\
    \midrule
    Cheetah-Jump    &   15          &   1152.98$\pm$120  &   \textbf{1721.25$\pm$56}   \\
    Cheetah-Crawl   &   -15         &   1239.00$\pm$57   &   \textbf{1348.19$\pm$274}  \\
    \bottomrule
    \specialrule{0em}{3pt}{3pt}
    \end{tabular}
    \caption{Returns on two offline tasks Cheetah-Jump and Cheetah-Crawl. Two datasets consist of 5 policies trained with base rewards and base rewards plus diversity rewards respectively.}
\end{table}
In Cheetah-Jump tasks, we relabel the data by adding an extra reward $c_{z}(z-z_{0})$ where $z_{0}$ is the initial position on the z axis, and $c_{z}$ is the coefficient of the extra reward. $c_{z}$ can either be positive or negative. For positive values of $c_{z}$, the cheetah is encouraged to jump while running forward. For negative values, crawling on the floor receives higher reward. We train 5 polices each for base rewards and base rewards plus diversity rewards, denoted by ``random'' and ``diverse'' respectively.
\subsection{Effects of the range of data distribution}
\begin{table}[ht]
    \centering
    \begin{tabular}{cccc}
    \toprule
    Angle     &   top 1       &   top 2 mixed &   all mixed   \\
    \midrule
    $0^{\circ}$   &   1236.26$\pm$247  &   \textbf{1437.24$\pm$31}   &   989.13$\pm$65   \\
    $60^{\circ}$    &   910.70$\pm$121  &   \textbf{1285.31$\pm$66} &   593.88$\pm$434  \\
    $120^{\circ}$   &   \textbf{1362.46$\pm$104}    &   917.10$\pm$218  &   281.88$\pm$301  \\
    $180^{\circ}$   &   829.65$\pm$139  &   \textbf{1034.41$\pm$224}    &   717.68$\pm$120  \\
    $240^{\circ}$   &   \textbf{1416.80$\pm$160}    &   1373.72$\pm$73  &   850.82$\pm$62   \\
    $300^{\circ}$   &   \textbf{1141.68$\pm$100}    &   1087.26$\pm$137 &   817.77$\pm$37   \\
    \bottomrule
    \specialrule{0em}{3pt}{3pt}
    \end{tabular}
    \caption{Returns on Ant-Angle tasks with different angles trained on different datasets. Top 1 dataset is the data buffer with highest return. Top 2 mixed dataset is a mixture of two highest-rewarded buffers. All mixed dataset is a mixture of all data buffers generated by unsupervisedly trained policies.}
    \label{tab:data}
\end{table}
With the help of Ant-Angle environment and the policies learned by unsupervised RL, we conduct several experiments to verify the conclusions from theoretical derivations. Apart from the data buffer with maximum return, we build a data buffer denoted by ``all mixed'', by mixing data generated by all 10 policies. We also mix the data from top 2 polices to create a ``top 2 mixed'' buffer.\par
Referring to the upper left figure in Figure \ref{fig:xytraj}, the ``top 2 mixed'' data buffer includes two policies lying on the left and the right side near direction $0^{\circ}$. the top two distributions have similar distance from the optimal distribution. Therefore the mixed distribution $\rho_{T}^{\pi^{\beta}}$ has similar distance to the optimal compared with the nearest distribution  $W_{1}(\rho_{T}^{\pi^{\beta}},\rho_{T}^{\pi^{*}})\approx\min_{i}W_{1}(\rho_{T}^{\pi_{i}^{\beta}},\rho_{T}^{*}))$. However, when all policies are mixed, it is obvious $W_{1}(\rho_{T}^{\pi^{\beta}},\rho_{T}^{\pi^{*}})>\min_{i}W_{1}(\rho_{T}^{\pi_{i}^{\beta}},\rho_{T}^{*}))$. According to Equation \ref{eqn9}, the top 2 mixed dataset will get higher return than all mixed dataset. Table \ref{tab:data} and results in Appendix D have verified this claim. From another aspect, the top 2 mixed data buffer has a wider distribution than the top 1 buffer. Therefore the top 2 mixed buffer has a larger radius $B_{\pi^{\beta}}$ which may worsen performance according to Equation \ref{eqn6}. Surprisingly, the top 2 mixed buffer makes higher return than top 1 single buffer. We can conjecture that $B_{\pi^{\beta}}$ plays an insignificant role in the lower bound and the approximation in Equation \ref{eqn8} and \ref{eqn9} is proper. In addition, the wide spread of the mixed data may improve the generalization ability of the transition model in MOPO, which contributes to the higher return than top 1 data buffer.
\section{Discussion}
\subsection{Limitations}
Our derivation is based on the continuous state space with assumption that the transition function and the value functions are Lipschitz. There are some tasks may break the assumptions, e.g., pixel based tasks like Atari, non-smooth reward functions in goal reaching tasks \cite{tassa2018deepmind}. Therefore, it is necessary to verify the feasibility of UDG on these tasks in future deployment. We also adopt a non-parametric transition model in derivation. In practical model-based offline RL approaches, neural models have greater generalization ability than the non-parametric model. The influence of data distribution on neural models is not addressed by this work. In addition, whether can UDG be generalized with model-free offline RL algorithms remains unclear. Another limitation is at the unsupervised training stage, the diversity reward is calculated on the low dimensional space where the reward function is defined, e.g., the x-y plane in Ant-Angle. This requires prior knowledge of how the reward is computed. Nevertheless, the limitations mentioned above indicate interesting further research.
\subsection{Societal impact}
The UDG framework contains a stage of unsupervised RL. At this stage, the agent is not provided with any reward for solving any task. During the process of training, the agent may unexpectedly exploit the states in unsafe regions. Especially when deployed in realistic environments, it could incur damage of the environment or the robotic agent itself, or cause injury in robot-human interactions. Any deployment of UDG in the real world should be carefully designed to avoid safety incidents.
\subsection{Conclusion}
In this study we propose a framework UDG addressing data generation issues in offline reinforcement learning. In order to solve unknown tasks at the offline training stage, UDG first employs unsupervised RL and obtains a series of diverse policies for data generation. The experience generated by each policy is relabeled according the reward function adopted before the offline training stage. The final step of UDG is to select the data buffer with highest average return and to feed the data to model-based offline RL algorithms like MOPO. We provide theoretical analysis on the performance gap between the offline learned policy and the optimal policy w.r.t the distribution of the batch data. We also reveal that UDG is an approximate minimal worst-case regret approach under the task-agnostic setting. Our experiments evaluate UDG on two locomotive tasks, Ant-Angle and Cheetah-Jump. Empirical results on multiple offline tasks demonstrate UDG is overall better than data generated by a policy dedicated to solve a specific task. Additional experiments show that the range of data distribution has minor effects on performance and the distance from the optimal policy is the most important factor. It is also confirmed that choosing the data buffer with highest return is necessary for better performance.
\medskip
{
\bibliography{refer}
}


\end{document}